\documentclass[11pt,a4paper]{article}
\usepackage[hyperref]{emnlp2019}
\usepackage{times}
\usepackage{latexsym} 
\usepackage[algoruled,boxed,noend]{algorithm2e}
\usepackage{amssymb}
\usepackage{amsmath}
\usepackage{url} 
\SetKwInOut{Parameter}{parameter}
\usepackage{enumitem}
\usepackage{mathtools}
\usepackage{fontawesome} 
\definecolor{cyan}{HTML}{38F7F1}
\definecolor{yellow}{HTML}{FDA062}
\usepackage{lipsum}
\setitemize{noitemsep,topsep=0pt,parsep=0pt,partopsep=0pt}
\let\oldnl\nl
\newcommand{\nonl}{\renewcommand{\nl}{\let\nl\oldnl}}

\SetAlFnt{\small}
\SetAlCapFnt{\large}
\SetAlCapNameFnt{\large}
\usepackage{algorithmic} 
\usepackage{calc}
\usepackage{pifont}
\newcommand{\cmark}{\ding{51}}%
\newcommand{\xmark}{\ding{55}}%
\algsetup{linenosize=\tiny}
\aclfinalcopy 
\newlength{\Width}%
\newlength{\DepthReference}
\newlength{\HeightReference}
\newcommand{\MyColorBox}[2][red]%
{%
    \settowidth{\Width}{#2}%
    \colorbox{#1}%
    {%
        \raisebox{-\DepthReference}%
        {%
                \parbox[b][\HeightReference+\DepthReference][c]{\Width}{\centering#2}%
        }%
    }%
}
\newcommand{\llIf}[2]{{\let\par\relax\lIf{#1}{#2}}}
\newcommand{\llElse}[1]{{\let\par\relax\lElse{#1}}} 
\usepackage[export]{adjustbox}
\usepackage{booktabs,array}
\SetKwFunction{KagNet}{\textsc{KagNet}} 

\newcommand{\secref}[1]{\S\ref{#1}}

\title{KagNet: Knowledge-Aware Graph Networks\\ for Commonsense Reasoning}

\author{
Bill Yuchen Lin\textsuperscript{\dag}~~~
Xinyue Chen\textsuperscript{\ddag}~~
Jamin Chen\textsuperscript{\dag}~~
Xiang Ren\textsuperscript{\dag}\\
\texttt{\{yuchen.lin,jaminche,xiangren\}@usc.edu},~~\texttt{kiwisher@sjtu.edu.cn}
\\~\\
\textsuperscript{\dag}{Computer Science Department, University of Southern California} \\ \textsuperscript{\ddag}Computer Science Department, Shanghai Jiao Tong University
}

\date{}

\begin{document}
\maketitle
\begin{abstract}

Commonsense reasoning aims to empower machines with the human ability to make presumptions about ordinary situations in our daily life.  
In this paper, we propose a textual inference framework for answering commonsense questions, which effectively utilizes external, structured commonsense knowledge graphs to perform \textit{explainable} inferences. 
The framework first grounds a question-answer pair from the semantic space to the knowledge-based symbolic space as a schema graph, a related sub-graph of external knowledge graphs.
It represents schema graphs with a novel knowledge-aware graph network module named \KagNet, and finally scores answers with graph representations.
Our model is based on graph convolutional networks and LSTMs, with a hierarchical path-based attention mechanism.
The intermediate attention scores make it transparent and interpretable, which thus produce trustworthy inferences.
Using \texttt{ConceptNet} as the only external resource for \textsc{Bert}-based models, we achieved state-of-the-art performance on the \texttt{CommonsenseQA}, a large-scale dataset for commonsense reasoning. We open-source our code\footnote{\url{https://github.com/INK-USC/KagNet}} to the community for future research in knowledge-aware commonsense reasoning.
\end{abstract}

\section{Introduction}
Human beings are rational 
and a major component of rationality is the ability to reason.
Reasoning is the process of combining facts and beliefs to make new decisions~\cite{johnson1980mental}, as well as the ability to manipulate knowledge to draw inferences~\cite{Hudson2018CompositionalAN}.
Commonsense reasoning utilizes the basic knowledge that reflects our natural understanding of the world and human behaviors, which is common to all humans.

Empowering machines with the ability to perform commonsense reasoning has been seen as the bottleneck of artificial general intelligence~\cite{davis2015commonsense}.
Recently, there have been a few emerging large-scale datasets for testing machine commonsense with various focuses~\cite{Zellers2018SWAGAL, Sap2019SocialIQACR, Zellers2019FromRT}.
In a typical dataset, \texttt{CommonsenseQA}~\cite{Talmor2018CommonsenseQAAQ},
given a question like \textit{``Where do adults use glue sticks?''}, with the answer choices being \{classroom(\xmark), {office} (\cmark), desk drawer (\xmark)\},
a commonsense reasoner is expected to differentiate the correct choice from other ``distractive'' candidates. 
False choices are usually highly related to the question context, but just less possible in real-world scenarios, making the task even more challenging.
This paper aims to tackle the research question of how we can teach machines to make such commonsense inferences, particularly in the question-answering setting.
\begin{figure}
	\centering
	\includegraphics[width=0.9\linewidth]{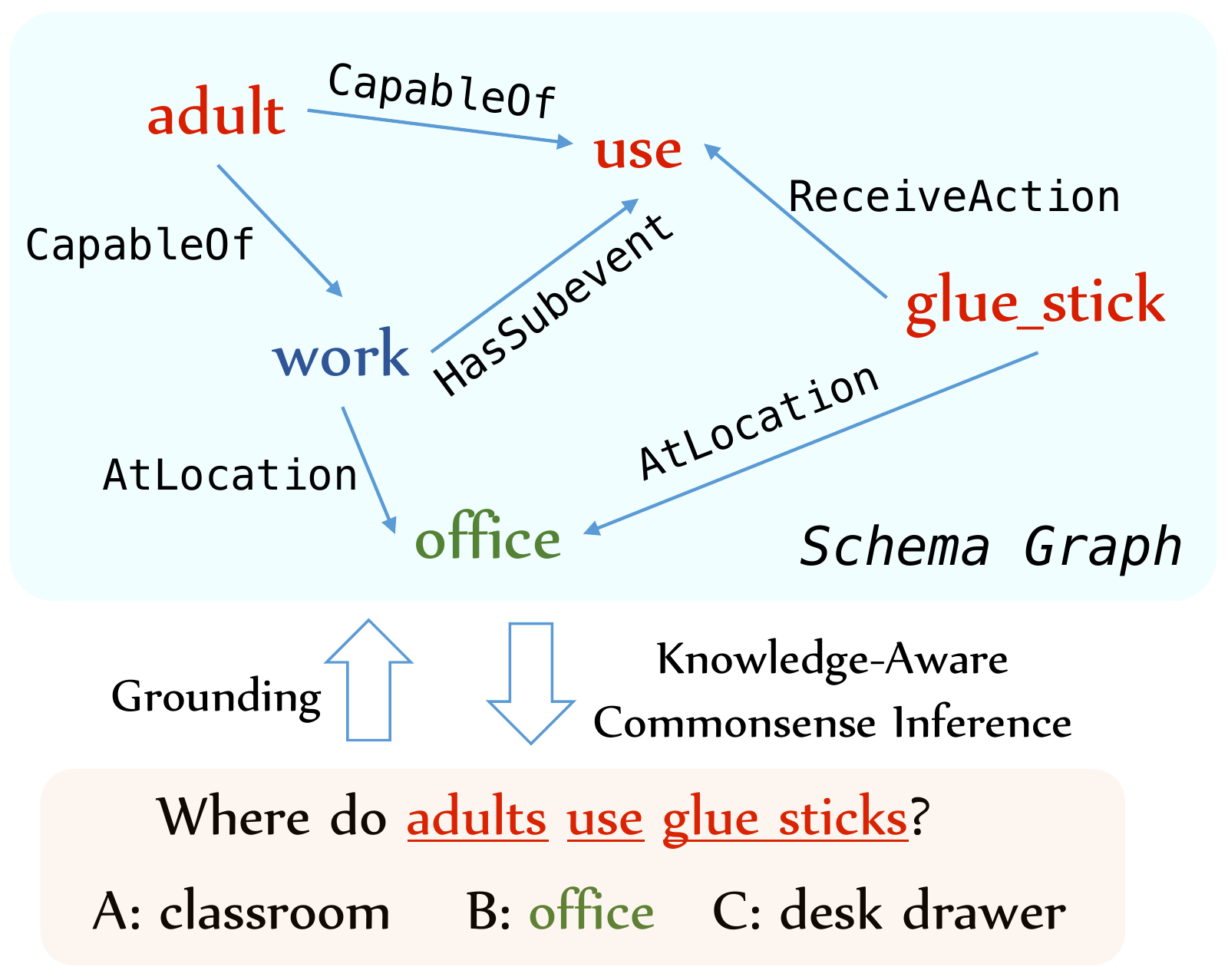}
	\caption{An example of using external commonsense knowledge (\MyColorBox[cyan!10]{symbolic space})
	for inference in natural language commonsense questions (\MyColorBox[yellow!10]{semantic space}). }
	\label{fig:example}
\end{figure}

It has been shown that simply fine-tuning large, pre-trained language models such as \textsc{Gpt}~\cite{radford2018improving} and \textsc{Bert}~\cite{Devlin2019} can be a very strong baseline method.
However, there still exists a large gap between performance of said baselines and human performance.
Reasoning with neural models is also lacking in transparency and interpretability.
There is no clear way as to how they manage to answer commonsense questions, thus making their inferences dubious.

Merely relying on pre-training large language models on corpora cannot provide well-defined or reusable structures for explainable commonsense reasoning. 
We argue that it would be more beneficial to propose reasoners that can exploit commonsense knowledge bases~\cite{Speer2017ConceptNet5A, Tandon2017WebChild2, sap2018atomic}. 
Knowledge-aware models can explicitly incorporate external knowledge as relational inductive biases~\cite{Battaglia2018RelationalIB} to enhance their reasoning capacity, as well as to increase the transparency of model behaviors for more interpretable results. 
Furthermore, a knowledge-centric approach is extensible through commonsense knowledge acquisition techniques~\cite{li2016commonsense, Xu2018AutomaticEO}.

We propose a knowledge-aware reasoning framework for learning to answer commonsense questions, which has two major steps: schema graph grounding~(\secref{sec:grounding}) and graph modeling for inference~(\secref{sec:kagnet}).
As shown in Fig.~\ref{fig:example}, for each pair of question and answer candidate, we retrieve a graph from external knowledge graphs (e.g. \texttt{ConceptNet}) in order to capture the relevant knowledge for determining the plausibility of a given answer choice.
The graphs are named ``schema graphs'' inspired by the \texttt{schema theory} proposed by Gestalt psychologists~\cite{axelrod1973schema}.
The grounded schema graphs are usually much more complicated and noisier, unlike the ideal case shown in the figure. 

Therefore, we propose a knowledge-aware graph network module to further effectively model schema graphs.  
Our model \KagNet is a combination of graph convolutional networks \cite{kipf2016semi} and LSTMs, with a hierarchical path-based attention mechanism, which forms a \texttt{GCN-LSTM-HPA} architecture for path-based relational graph representation. 
Experiments show that our framework achieved a new state-of-the-art performance\footnote{The highest score on the leaderboard as of the time when we submitted the paper (May 2019).} on the \texttt{CommonsenseQA} dataset.
Our model also works better then other methods with limited supervision, and provides human-readable results via intermediate attention scores.


\section{Overview}
\label{sec:overview}
In this section, we first formalize the commonsense question answering  problem in a knowledge-aware setting, and then introduce the overall workflow of our framework. 
\begin{figure}
	\centering
	\includegraphics[width=1\linewidth]{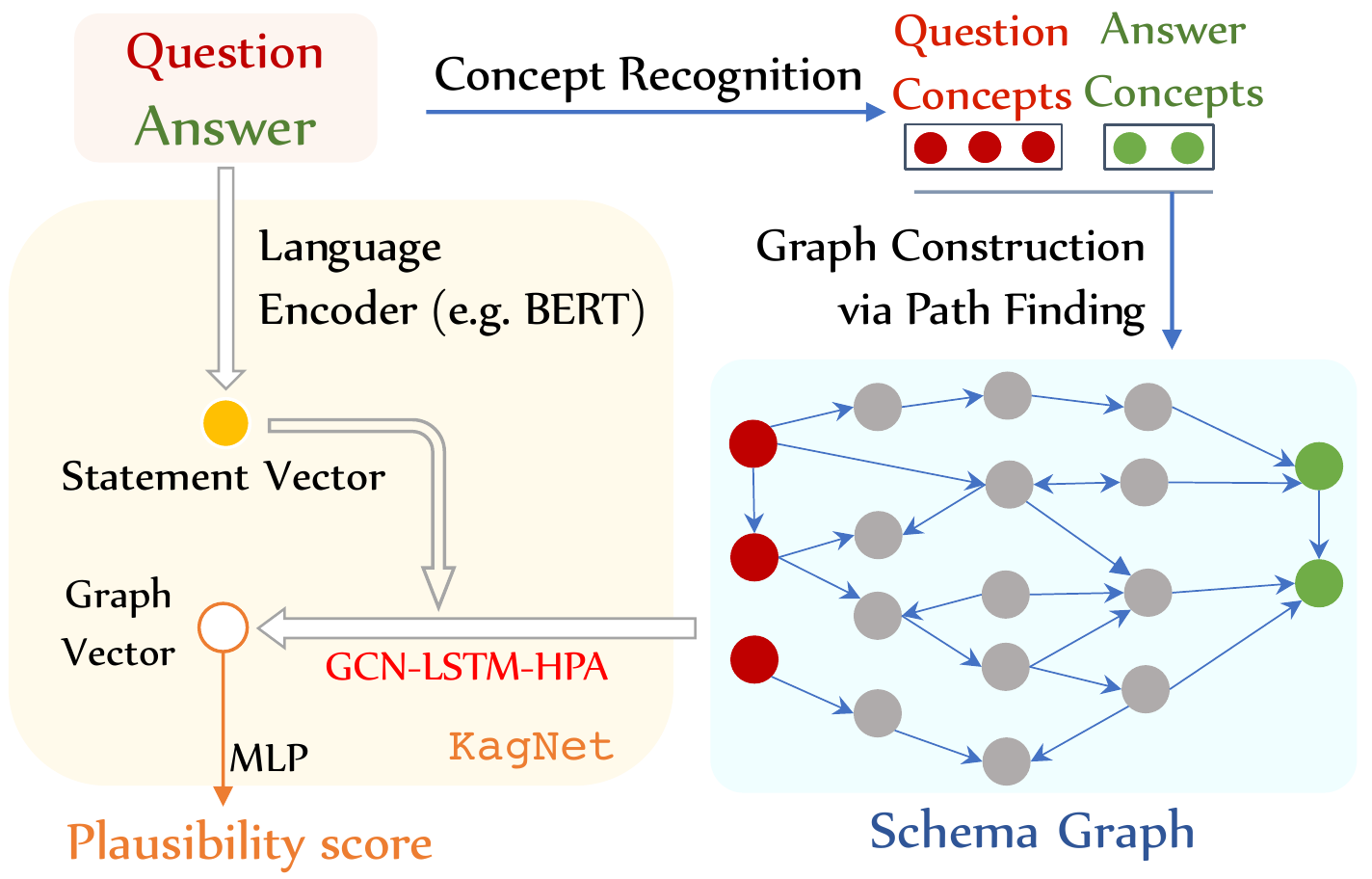}
	\caption{The overall workflow of the proposed framework with knowledge-aware graph network module.\vspace{0pt}}
	\label{fig:overviewlatex}
\end{figure}
\subsection{Problem statement}
Given a commonsense-required natural language question $q$ and a set of $N$ candidate answers $\{a_i\}$, 
the task is to choose one answer from the set. 
From a knowledge-aware perspective, we additionally assume that the question $q$ and choices $\{a_i\}$ can be grounded as a \textbf{schema graph} (denoted as $g$) extracted from a large external {knowledge graph} $G$, which is helpful for measuring the plausibility of answer candidates.
The knowledge graph $G=(V,E)$ can be defined as a fixed set of \textbf{concepts} $V$, and \textbf{typed edges} $E$ describing semantic relations between concepts. 
Therefore, our goal is to effectively ground and model schema graphs to improve the reasoning process.

\subsection{Reasoning Workflow}
As shown in Fig.~\ref{fig:overviewlatex}, our framework accepts 
a pair of question and answer (QA-pair) denoted as $q$ and $a$.
It first recognizes  the mentioned concepts within them respectively from the concept set $V$ of the knowledge graph.
We then algorithmically construct the schema graph $g$ by finding paths between pairs of mentioned concepts~(\secref{sec:grounding}).

The grounded schema graph is further encoded with our proposed knowledge-aware graph network module~(\secref{sec:kagnet}).
We first use a model-agnostic language encoder, which can either be trainable or a fixed feature extractor, to represent the QA-pair as a statement vector.
The statement vector serves as an additional input to a \texttt{GCN-LSTM-HPA} architecture for path-based attentive graph modeling to obtain a graph vector. 
The graph vector is finally fed into a simple multi-layer perceptron to score this QA-pair into a scalar ranging from $0$ to $1$, representing the plausibility of the inference.
The answer candidate with the maximum plausibility score to the same question becomes the final choice of our framework.

\section{Schema Graph Grounding}
\label{sec:grounding}
The grounding stage is three-fold: recognizing concepts mentioned in text (\secref{sec:mcr}), constructing schema graphs by retrieving paths in the knowledge graph (\secref{sec:graphcreate}), and pruning noisy paths (\secref{sec:pruning}).

\subsection{Concept Recognition}
\label{sec:mcr}
We match tokens in questions and answers to sets of mentioned concepts ($\mathcal{C}_q$ and $\mathcal{C}_a$ respectively)  from the knowledge graph $G$ (for this paper we chose to use \texttt{ConceptNet} due to its generality).

A naive approach to mentioned concept recognition is 
to exactly match n-grams in sentences with the surface tokens of concepts in $V$.
For example, in the question ``\textit{Sitting too close to watch tv can cause what sort of pain?}'', the exact matching result $\mathcal{C}_q$ would be \{sitting, close, watch\_tv, watch, tv, sort, pain, etc.\}. 
We are aware of the fact that such retrieved mentioned concepts are not always perfect
(e.g. ``sort'' is not a semantically related concept, ``close'' is a polysemous concept).
How to efficiently retrieve contextually-related knowledge from noisy knowledge resources is still an open research question by itself~\cite{weissenborn2017dynamic, Khashabi2017LearningWI}, and thus most prior works choose to stop here~\cite{Zhong2018ImprovingQA, Wang2019ImprovingNL}.
We enhance this straightforward approach with some rules, such as soft matching with lemmatization and filtering of stop words, and further deal with noise by pruning paths (\secref{sec:pruning}) and reducing their importance with attention mechanisms (\secref{sec:att}).
%

%

\subsection{Schema Graph Construction}
\label{sec:graphcreate}
\textbf{ConceptNet.} Before diving into the construction of schema graphs, we would like to briefly introduce our target knowledge graph \texttt{ConceptNet}.
\texttt{ConceptNet} can be seen as a large set of triples of the form $(h, r, t)$, like (\texttt{ice}, \texttt{HasProperty}, \texttt{cold}), where $h$ and $t$ represent head and tail concepts in the concept set $V$ and $r$ is a certain relation type from the pre-defined set $R$. 
We delete and merge the original 42 relation types into 17 types, in order to increase the density of the knowledge graph\footnote{The full mapping list is in the \textsf{appendix}.} for grounding and modeling.

\textbf{Sub-graph Matching via Path Finding.}
We define a schema graph as a sub-graph $g$ of the whole knowledge graph $G$, which represents the related knowledge for reasoning a given question-answer pair with minimal additional concepts and edges.
One may want to find a minimal spanning sub-graph covering all the question and answer concepts, 
which is actually the  NP-complete ``Steiner tree problem'' in graphs~\cite{garey1977rectilinear}.
Due to the incompleteness and tremendous size of \texttt{ConceptNet}, we find that it is impractical to retrieve a comprehensive but helpful set of knowledge facts this way.
Therefore, 
we propose a straightforward yet effective graph construction algorithm via path finding among mentioned concepts ($\mathcal{C}_q \cup \mathcal{C}_a$).

Specifically, for each question concept $c_i \in \mathcal{C}_q$ and answer concept $c_j \in \mathcal{C}_a$, we can efficiently find paths between them that are shorter than $k$ concepts\footnote{We set $k=4$ in experiments to gather three-hop paths.}.
Then, we add edges, if any, between the concept pairs within
$\mathcal{C}_q$  or $\mathcal{C}_a$.

\subsection{Path Pruning via KG Embedding}
\label{sec:pruning}
To prune irrelevant paths from potentially noisy schema graphs, 
we first utilize knowledge graph embedding (KGE) techniques, like TransE~\cite{Wang2014KnowledgeGE}, to pre-train concept embeddings $\mathbf{V}$ and relation type embeddings $\mathbf{R}$, which are also used as initialization for~\KagNet (\secref{sec:kagnet}).
In order to measure the quality of a path, we decompose it into a set of triples, the confidence of which can be directly measured by the scoring function of the KGE method (i.e. the confidence of triple classification).
Thus, we score a path with the multiplication product of the scores of each triple in the path, and then we empirically set a threshold for pruning (\secref{sec:detail}).  
 
 \begin{figure*}
	\centering
	\includegraphics[width=1\linewidth]{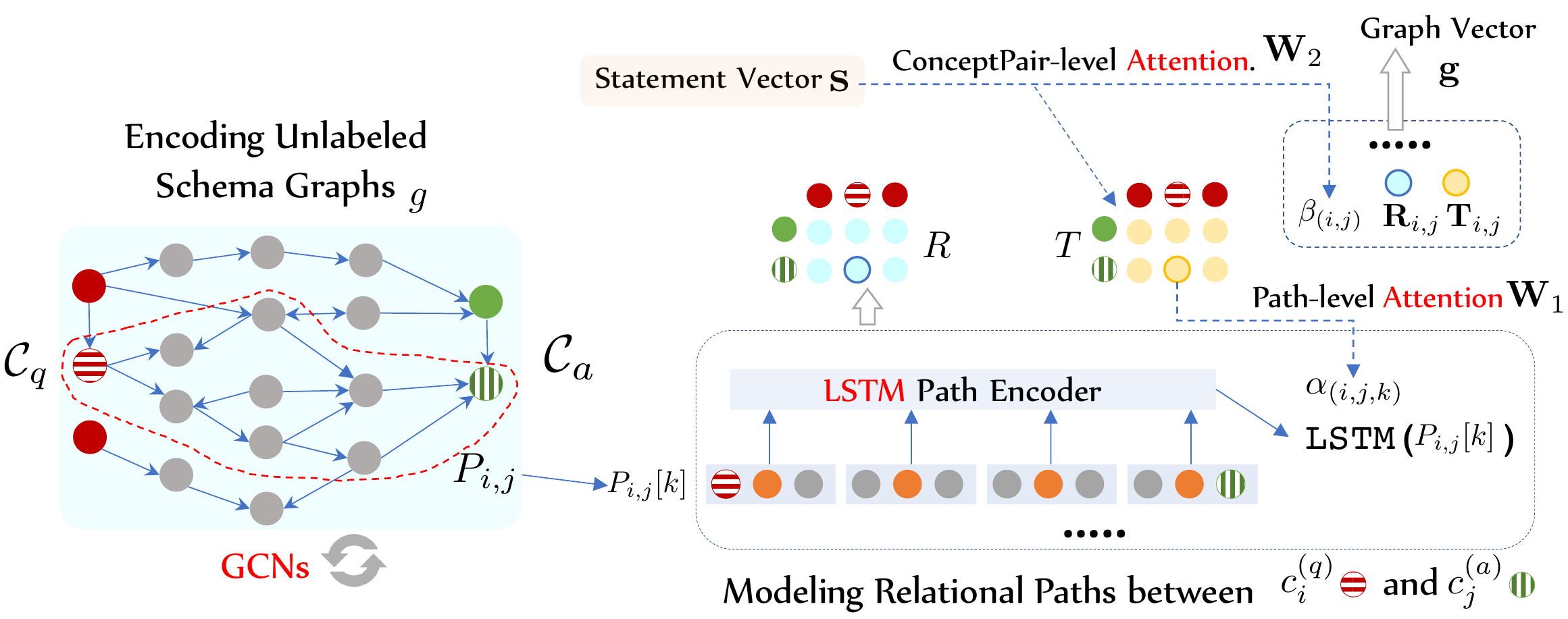}
	\caption{Illustration of the \texttt{GCN-LSTM-HPA} architecture for the proposed \KagNet module.}
	\label{fig:kagnet}
\end{figure*}

\section{Knowledge-Aware Graph Network}
\label{sec:kagnet}
The core component of our reasoning framework is the knowledge-aware graph network module \KagNet.
The \KagNet first encodes plain structures of schema graphs with graph convolutional networks (\secref{sec:gcn}) to accommodate pre-trained concept embeddings in their particular context within schema graphs.
It then utilizes LSTMs to encode the paths between $\mathcal{C}_q$ and $\mathcal{C}_a$, 
capturing multi-hop relational information (\secref{sec:lstm}).
Finally, we apply a hierarchical path-based attention mechanism (\secref{sec:att}) to complete the \texttt{GCN-LSTM-HPA} architecture, which models relational schema graphs with respect to the paths between question and answer concepts.
\subsection{Graph Convolutional Networks}
\label{sec:gcn}
Graph convolutional networks (GCNs) encode graph-structured data by updating node vectors via pooling features of their adjacent nodes~\cite{kipf2016semi}.
Our intuition for applying GCNs to schema graphs is to 1) contextually refine the concept vectors and 2) capture structural patterns of schema graphs for generalization.

Although we have obtained concept vectors by pre-training (\secref{sec:pruning}), 
the representations of concepts still need to be further accommodated to their specific schema graphs context.
Think of polysemous concepts such as ``close'' (\secref{sec:mcr}), which can either be a verb concept like in ``close the door'' or an adjective concept meaning ``a short distance apart''.
Using GCNs to update the concept vector with their neighbors is thus helpful for disambiguation and contextualized concept embedding. 
Also, the pattern of schema graph structures provides potentially valuable information for reasoning. 
For instance, shorter and denser connections between question and answer concepts could mean higher plausibility under specific contexts. 

As many works show~\cite{Marcheggiani2017EncodingSW,Zhang2018GraphCO}, relational GCNs ~\cite{Schlichtkrull2018ModelingRD} usually over-parameterize 
the model and cannot effectively utilize multi-hop relational information.
We thus apply GCNs on the plain version (unlabeled, non-directional) of schema graphs, ignoring relation types on the edges.
Specifically,
the vector for concept $c_i\in \mathcal{V}_g$ in the schema graph $g$ is initialized by their pre-trained embeddings at first ($h_i^{(0)} = \mathbf{V}_i$).
Then, we update them at the $(l+1)$-th layer by pooling features of 
their neighboring nodes ($N_i$) and their own at the  $l$-th layer 
with an non-linear activation function $\sigma$:
\begin{align*} 
h_i^{(l+1)} = \sigma (W_{self}^{(l)}h_i^{(l)}+\sum_{j\in N_i}\frac{1}{|N_i|}W^{(l)}h_j^{(l)})
\end{align*}

\subsection{Relational Path Encoding}
\label{sec:lstm}
In order to capture the relational information in schema graphs, we propose an LSTM-based path encoder on top of the outputs of GCNs. 
Recall that our graph representation has a special purpose: ``\textit{to measure the plausibility of a candidate answer to a given question}''. 
Thus, we propose to represent graphs with respect to the paths between question concepts $\mathcal{C}_q$ and answer concepts $\mathcal{C}_a$. 

We denote the $k$-th path between $i$-th question concept $c_i^{(q)}\in \mathcal{C}_q$ and $j$-th answer concept $c_j^{(a)}\in \mathcal{C}_a$ as $P_{i,j}[k]$, which is a sequence of triples: 
\begin{align*}
	P_{i,j}[k] = [(c_i^{(q)}, r_0, t_0),...,(t_{n-1}, r_n,c_j^{(a)} )]
\end{align*}

Note that the relations are represented with trainable relation vectors (initialized with pre-trained relation embeddings), and concept vectors are the GCNs' outputs ($h^{(l)}$).
Thus, each triple can be represented by the concatenation of the three corresponding vectors. 
We employ LSTM networks to encode these paths as sequences of triple vectors, taking the concatenation of the first and the last hidden states:
$$\vspace{-10pt} \mathbf{R}_{i,j}= \frac{1}{|P_{i,j}|}\sum_k \texttt{LSTM}(P_{i,j}[k]) $$

The above $\mathbf{R}_{i,j}$ can be viewed as the latent relation between the question concept $c_i^{(q)}$ and the answer concept $c_j^{(a)}$, for which we aggregate the representations of all the paths between them in the schema graph.
Now we can finalize the vector representation of a schema graph $\mathbf{g}$ by aggregating all vectors in the matrix $\mathbf{R}$ using mean pooling:
\begin{align*}
    \mathbf{T}_{i,j} &= \texttt{MLP}([\mathbf{s}~;~ \mathbf{c_q^{(i)}}~;~ \mathbf{c_a^{(j)}}]) \\
    \mathbf{g}&= \frac{\sum_{i,j}   [\mathbf{R}_{i,j}~;~ \mathbf{T}_{i,j}] }{|\mathcal{C}_q|\times|\mathcal{C}_a|}
\end{align*}
, where $[\cdot~;~\cdot]$ means concatenation of two vectors.

The statement vector $\mathbf{s}$ in the above equation is obtained from a certain language encoder, which can either be a trainable sequence encoder like LSTM or features extracted from pre-trained universal language encoders like \textsc{Gpt}/\textsc{Bert}). {To encode a question-answer pair with universal language encoders,
we simply create a  sentence combining the question and the answer with a special token (``\texttt{question}+ [sep] + \texttt{answer}''), and then use the vector of `[cls]' as suggested by prior works~\cite{Talmor2018CommonsenseQAAQ}.}.

We concatenate \MyColorBox[cyan!10]{$\mathbf{R}_{i,j}$} with an additional vector \MyColorBox[yellow!10]{$\mathbf{T}_{i,j}$} before doing average pooling.
The $\mathbf{T}_{i,j}$ is inspired from the \texttt{Relation Network}~\cite{Santoro2017ASN}, which also encodes the latent relational information yet from the context in the statement ${s}$ instead of the schema graph $g$. 
Simply put, we want to combine the relational representations of a pair of question/answer concepts from both the schema graph side (\MyColorBox[cyan!10]{symbolic space}) and the language side (\MyColorBox[yellow!10]{semantic space}).

Finally,
the plausibility score of the answer candidate $a$ to the question $q$ can be computed as $\texttt{score}(q,a) = \texttt{sigmoid}(\texttt{MLP}(\mathbf{g}))$.

\subsection{Hierarchical Attention Mechanism}
\label{sec:att}  
A natural argument against the above \texttt{GCN-LSTM-mean} architecture is that 
mean pooling over the path vectors does not always make sense, since some paths are more important than others for reasoning.
Also, it is usually not the case that all pairs of question and answer concepts equally contribute to the reasoning.  
Therefore, we propose a hierarchical path-based attention mechanism to selectively aggregate important path vectors and then more important question-answer concept pairs. 
This core idea is similar to the work of \citeauthor{Yang2016HierarchicalAN} (2016),
which proposes a document encoder that has two levels of attention mechanisms applied at the word- and sentence-level.
In our case, we have path-level and concept-pair-level attention for learning to contextually model graph representations. 
We learn a parameter matrix $\mathbf{W}_1$ for path-level attention scores, and the importance of the path $P_{i,j}[k]$ is denoted as $\hat{\alpha}_{(i,j,\cdot)}$.  
\begin{align*}
	\alpha_{(i,j,k)} &= \mathbf{T}_{i,j} ~\mathbf{W}_{1} ~\texttt{LSTM}(P_{i,j}[k]) ,\\
	\hat{\alpha}_{(i,j,\cdot)} &= 	\texttt{SoftMax}(\alpha_{(i,j,\cdot)}),\\
	\hat{\mathbf{R}}_{i,j} &= \sum_k \hat{\alpha}_{(i,j,k)} \cdot \texttt{LSTM}(P_{i,j}[k]).  
\end{align*} 
Afterwards, we similarly obtain the attention over concept-pairs. 
\begin{align*} 
	\beta_{(i,j)} &=  \mathbf{s_{}}~\mathbf{W}_{2} ~ {\mathbf{T}}_{i,j}  \\
	\hat{\beta}_{(\cdot,\cdot)} &= \texttt{SoftMax}(\beta_{(\cdot,\cdot)})\\
	\hat{\mathbf{g}} &= \sum_{i,j} \hat{\beta}_{(i,j)} [\hat{\mathbf{R}}_{i,j}~;~ \mathbf{T}_{i,j}] 
\end{align*} 

The whole \texttt{GCN-LSTM-HPA} architecture is illustrated in Figure~\ref{fig:kagnet}.
To sum up, we claim that the \KagNet is a graph neural network module with the \texttt{GCN-LSTM-HPA} architecture that models relational graphs for relational reasoning under the context of both \MyColorBox[cyan!10]{knowledge symbolic space} and  \MyColorBox[yellow!10]{language semantic space}. 

\section{Experiments}
\label{sec:exp}
We introduce our setups of the \texttt{CommonsenseQA} dataset~\cite{Talmor2018CommonsenseQAAQ}, present the baseline methods, and finally analyze experimental results.

\begin{table*}[t]
	\centering
	\scalebox{0.72
	}{
		\begin{tabular}{@{}lccccccccc@{}}
			\toprule   
			 &    
			\multicolumn{2}{c}{\textbf{10}(\%) of IHtrain}  &  
			\multicolumn{2}{c}{\textbf{50}(\%) of IHtrain}  &  
			\multicolumn{2}{c}{\textbf{100}(\%) of IHtrain}  &  
			\\
			
			\textbf{Model}&    
			\multicolumn{1}{c}{IHdev-Acc.(\%)}  &    \multicolumn{1}{c}{IHtest-Acc.(\%)}  &
			\multicolumn{1}{c}{IHdev-Acc.(\%)}  &    \multicolumn{1}{c}{IHtest-Acc.(\%)}  &
			\multicolumn{1}{c}{IHdev-Acc.(\%)}  &    \multicolumn{1}{c}{IHtest-Acc.(\%)}  &
			
			\\
			\midrule 
			
			Random guess & 20.0& 20.0& 20.0& 20.0& 20.0& 20.0\\
			\midrule \midrule 
			\textsc{Gpt-FineTuning} & 27.55 & 26.51 & 32.46 & 31.28& 47.35 & 45.58  \\
			\textsc{Gpt-\textbf{KagNet}} & 28.13 & \textbf{26.98 } &  33.72 & \textbf{32.33} & 48.95  & \textbf{46.79}\\  
			
			\midrule \midrule 
			\textsc{Bert-Base-FineTuning} & 30.11 & 29.78 & 38.66 & 36.83 & 53.48 & 53.26  \\
			\textsc{Bert-Base-\textbf{KagNet}} & 31.05 & \textbf{30.94} & 40.32 & \textbf{39.01} & 55.57 & \textbf{56.19}  \\  
			\midrule
			\textsc{Bert-Large-FineTuning} & 35.71 & 32.88 & 55.45 & 49.88 & 60.61 & 55.84  \\
			\textsc{Bert-Large-\textbf{KagNet}} & 36.82  & \textbf{33.91}  & 58.73 & \textbf{51.13} & 62.35 & \textbf{57.16}  \\  
			\midrule\midrule
			Human Performance & - & \textit{88.9}& - & \textit{88.9}& - & \textit{88.9}  \\
			\bottomrule
		\end{tabular}
	} 
	\caption{Comparisons with large pre-trained language model fine-tuning with different amount of training data. \vspace{0pt}}
	\label{tab:IHfrac}
\end{table*}
\subsection{Dataset and Experiment Setup}
The \texttt{CommonsenseQA} dataset consists of 12,102 (v1.11) natural language questions in total that require human commonsense reasoning ability to answer, where each question has five candidate answers (hard mode).
The authors also release an easy version of the dataset by picking two random terms/phrases for sanity check.
\texttt{CommonsenseQA} is directly gathered from real human annotators and covers a broad range of types of commonsense, including spatial, social, causal, physical, temporal, etc.
To the best of our knowledge, \texttt{CommonsenseQA} may be the most suitable choice
for us to evaluate supervised learning models for question answering.

For the comparisons with the reported results in the \texttt{CommonsenseQA}'s paper and leaderboard, 
we use the official split (9,741/1,221/1,140) named (OFtrain/OFdev/OFtest).
Note that 
the performance on OFtest can only be tested
by submitting predictions to the organizers.
To efficiently test other baseline methods and ablation studies, 
we choose to use randomly selected 1,241 examples from the training data as our in-house data, 
forming an (8,500/1,221/1,241) split denoted as (IHtrain/IHdev/IHtest).
All experiments are using the random-split setting as the authors suggested, and three or more random states are tested on development sets to pick the best-performing one.

\subsection{Compared Methods}
We consider two different kinds of baseline methods as follows:

\smallskip
\noindent
$\bullet$ \textit{Knowledge-agnostic Methods.} 
These methods either use no external resources or only use unstructured textual corpora as additional information, including gathering textual snippets from search engine or large pre-trained language models like \textsc{Bert-Large}. 
\textsc{QABilinear}, \textsc{QACompare}, \textsc{ESIM} are three supervised learning models for natural language inference that can be equipped with different word embeddings including \texttt{GloVe} and \textsc{ELMo}.
\texttt{BIDAF++} utilizes Google web snippets as context 
and is further
augmented with a self-attention layer 
while using \textsc{ELMo} as input features.
\textsc{Gpt}/\textsc{Bert-Large} are fine-tuning methods with an additional linear layer for classification as the authors suggested. 
They both add a special token `[sep]' to
the input and use the hidden state of the `[cls]' as the input to the linear layer.
More details about them can be found in the dataset paper~\cite{Talmor2018CommonsenseQAAQ}.

\smallskip
\noindent
$\bullet$ \textit{Knowledge-aware Methods.}  
We also \textit{adopt} some recently proposed methods of incorporating knowledge graphs for question answering.
\textsc{KV-Mem}~\cite{Mihaylov2018KnowledgeableRE} is a method that incorporates retrieved triples from \texttt{ConceptNet} at the word-level, which uses a key-valued memory module to improve the representation of each token individually by learning an attentive aggregation of related triple vectors.
\textsc{CBPT}~\cite{Zhong2018ImprovingQA} is a plug-in method of assembling the predictions of any models with a straightforward method of utilizing pre-trained concept embeddings from \texttt{ConceptNet}.
\textsc{TextGraphCat}~\cite{Wang2018ImprovingNL} concatenates the graph-based and  text-based representations of the statement and then feed it into a classifier. 
We create sentence template for generating sentences and then feed retrieved triples as additional text inputs as a baseline method \textsc{TripleString}. 
\citeauthor{Rajani2019ExplainYL} (2019) propose to collect human explanations for commonsense reasoning from annotators as additional knowledge (\textsc{CoS-E}), and then train a language model based on such human annotations for improving the model performance.  

\begin{table}[t]

	\centering
	\scalebox{0.70
	}{
		\begin{tabular}{@{}lccccccccc@{}}
			\toprule   
			\textbf{Model}&    \multicolumn{1}{c}{OFdev-Acc.(\%)}  &    \multicolumn{1}{c}{OFtest-Acc.(\%)}  \\
			\midrule 
			 
			Random guess & 20.0& 20.0\\
			\midrule 
			\textsc{BIDAF++} & - & 32.0   \\
			\textsc{QACompare+Glove} & - & 25.7   \\
			\textsc{QABLinear+Glove} & - & 31.5   \\
			\textsc{Esim+Elmo} & - & 32.8   \\
			\textsc{Esim+Glove}  & - & 34.1   \\

			\midrule \midrule
			\textsc{Gpt-FineTuning} & 47.11 &$45.5$   \\
			\textsc{Bert-Base-FineTuning}  & 53.57 &$53.0$  \\
			\textsc{Bert-Large-FineTuning}  & 62.34 &$56.7$  \\
			\textsc{CoS-E} (w/ additional annotations)  & - &$58.2$  \\
			\midrule
			\textbf{\textsc{KagNet}} (Ours)  & 64.46 &\textbf{58.9}  \\
			\midrule\midrule
			Human Performance & - & \textit{88.9}  \\
			\bottomrule
		\end{tabular}
	} 

\caption{Comparison with official benchmark baseline methods using the official split on the leaderboard. \vspace{0pt}}
\label{tab:OFres}
\end{table}
\subsection{Implementation Details of KagNet}
\label{sec:detail}
Our best (tested on OFdev) settings of \KagNet have two GCN layers (100 dim, 50dim respectively), and one bidirectional LSTMs (128dim)
. We pre-train KGE
using TransE (100 dimension) initialized with GloVe embeddings. 
The statement encoder in use is \textsc{Bert-Large}, which works as a pre-trained sentence encoder to obtain fixed features for each pair of question and answer candidate.
The paths are pruned with path-score threshold set to \textit{0.15}, keeping \textit{67.21\%} of the original paths.
We did not conduct pruning on concept pairs with less than three paths. 
For very few pairs with none path, $\hat{\mathbf{R}}_{(i,j)}$ will be a randomly sampled vector. 
We learn our \KagNet models with Adam optimizers~\cite{Kingma2015AdamAM}.
In our experiments, we found that the recall of ConceptNet on commonsense questions and answers is very high (over 98\% of QA-pairs have more than one grounded concepts). 
\subsection{Performance Comparisons and Analysis}

\noindent
\textbf{Comparison with standard baselines.} \\
As shown in Table~\ref{tab:OFres}, we first use the official split to compare our model with the baseline methods reported on the paper and leaderboard. 
\textsc{Bert} and \textsc{Gpt}-based pre-training methods are much higher than other baseline methods, demonstrating the ability of language models to store commonsense knowledge in an implicit way.
This presumption is also investigated by~\citeauthor{trinh2019do} (2019) and ~\citeauthor{Wang2019DoesIM} (2019).
Our proposed framework achieves an absolute increment of  2.2\% in accuracy on the test data, a state-of-the-art
performance.

We conduct the experiments with our in-house splits to investigate whether our \KagNet can also work well on other universal language encoders (\textsc{GPT} and \textsc{Bert-Base}), particularly with different fractions of the dataset (say 10\%, 50\%, 100\% of the training data).
Table~\ref{tab:IHfrac} shows that our \KagNet-based methods using fixed pre-trained language encoders outperform fine-tuning themselves in all settings.
Furthermore, we find that the improvements in a small data situation (10\%) is relatively limited, and we believe an important future research direction is thus few-shot learning for commonsense reasoning.

\begin{table}[t]
	\centering
	\scalebox{0.63
	}{
		\begin{tabular}{@{}lccccccccc@{}}
			\toprule   
			&      
			\multicolumn{2}{c}{Easy Mode}  &  
			\multicolumn{2}{c}{Hard Mode}  &  
			\\
			
			\textbf{Model}&     
			\multicolumn{1}{c}{IHdev.(\%)}  &    \multicolumn{1}{c}{IHtest.(\%)}  &
			\multicolumn{1}{c}{IHdev.(\%)}  &    \multicolumn{1}{c}{IHtest.(\%)}   
			\\ 
			
			Random guess &   33.3& 33.3& 20.0& 20.0\\
			\midrule \midrule 
			\textsc{BLSTMs} &   80.15 & 78.01& 34.79 &32.12  \\
			\textsc{+ KV-MN} &  81.71 & 79.63& 35.70 & 33.43\\  
			 
			\textsc{+ CSPT} &   81.79 & 80.01 & 35.31 & 33.61  \\
			\textsc{+ TextGraphCat} & 82.68  & 81.03 & 34.72 & 33.15  \\   
			\textsc{+ TripleString} &   79.11 & 76.02 & 33.19 & 31.02 \\
			\textsc{+ \textbf{KagNet}} &  83.26 & \textbf{82.15} & 36.38 & \textbf{34.57}  \\  
			\midrule\midrule
			Human Performance &  - & \textit{99.5}& - & \textit{88.9}  \\
			\bottomrule
		\end{tabular}
	} 
	\caption{Comparisons with knowledge-aware baseline methods using the in-house split (both easy and hard mode) on top of BLSTM as the sentence encoder. }
	\label{tab:knowres}
\end{table}
\smallskip
\noindent
 \textbf{Comparison with knowledge-aware baselines.} \\
To compare our model with other adopted baseline methods that also incorporate \texttt{ConceptNet}, 
we set up a bidirectional LSTM networks-based model for our in-house dataset.
Then, we add baseline methods and \KagNet onto the BLSTMs to compare their abilities to utilize external knowledge\footnote{We do LSTM-based setup because it is non-trivial to apply token-level knowledge-aware baseline methods for complicated pre-trained encoders like \textsc{Bert}.}.
Table~\ref{tab:knowres} shows the comparisons under both easy mode and hard mode, and our methods outperform all knowledge-aware baseline methods by a large margin in terms of accuracy.
Note that we compare our model and the \texttt{CoS-E} in Table~\ref{tab:OFres}. 
Although \texttt{CoS-E} also achieves better result than only fine-tuning BERT by training with human-generated explanations, we argue that our proposed \texttt{KagNet} does not utilize any additional human efforts to provide more supervision. 

\smallskip
\noindent
\textbf{Ablation study on model components.} \\
To better understand the effectiveness of each component of our method, we have done ablation study as shown in Table~\ref{tab:abl}.
We find that replacing our GCN-LSTM-HPA architecture with traditional relational GCNs, which uses separate weight matrices for different relation types, results in worse performance, due to its over-parameterization.
The attention mechanisms matters almost equally in two levels, and pruning also effectively filters noisy paths.

\smallskip
\noindent
\textbf{Error analysis.} \\
In the failed cases, there are three kinds of hard problems that \KagNet is still not good at.
\begin{itemize}
    \item \textbf{negative reasoning:} the grounding stage is not sensitive to the negation words, and thus can choose exactly opposite answers.
    \item \textbf{comparative reasoning strategy:} 
    For the questions with more than one highly plausible answers, the commonsense reasoner should benefit from explicitly investigating the difference between different answer candidates, while \KagNet training method is not capable of doing so.
    
    \item \textbf{subjective reasoning:} Many answers actually depend on the ``personality'' of the reasoner. 
    For instance, ``\textit{Traveling from new place to new place is likely to be what?}'' 
    The dataset gives the answer as ``exhilarating'' instead of ``exhausting'', which we think is more like a personalized subjective inference instead of common sense.
    
\end{itemize}

\begin{table}[t]
	\centering
	\scalebox{0.75
	}{
		\begin{tabular}{@{}lccccccccc@{}}
			\toprule    
			\textbf{Model}&     
			\multicolumn{1}{c}{IHdev.(\%)}  &    \multicolumn{1}{c}{IHtest.(\%)}    
			\\  
			\midrule  
			\textsc{KagNet (standard)} &   62.35 & 57.16  \\
			\text{~: replace GCN-HPA-LSTM w/ R-GCN} &  60.01 & 55.08\\   
			\text{~: w/o GCN } &   61.84 & 56.11 \\
			\text{~: \#GCN Layers = 1} & 62.05  & 57.03   \\   
			\text{~: w/o Path-level Attention} &   60.12 &	56.05  \\
			\text{~: w/o QAPair-level Attention} &  60.39 &	56.13  \\  
			\text{~: using all paths (w/o pruning)} &  59.96 &	55.27\\  
			\bottomrule
		\end{tabular}
	} 
	\caption{Ablation study on the \KagNet framework. }
	\label{tab:abl}
\end{table}

\subsection{Case Study on Interpretibility} 
Our framework enjoys the merit of being more transparent, and thus provides more interpretable inference process.
We can understand our model behaviors by analyzing the hierarchical attention scores on the question-answer concept pairs and path between them. 

\begin{figure}[t]
	\centering
	\includegraphics[width=1.0\linewidth]{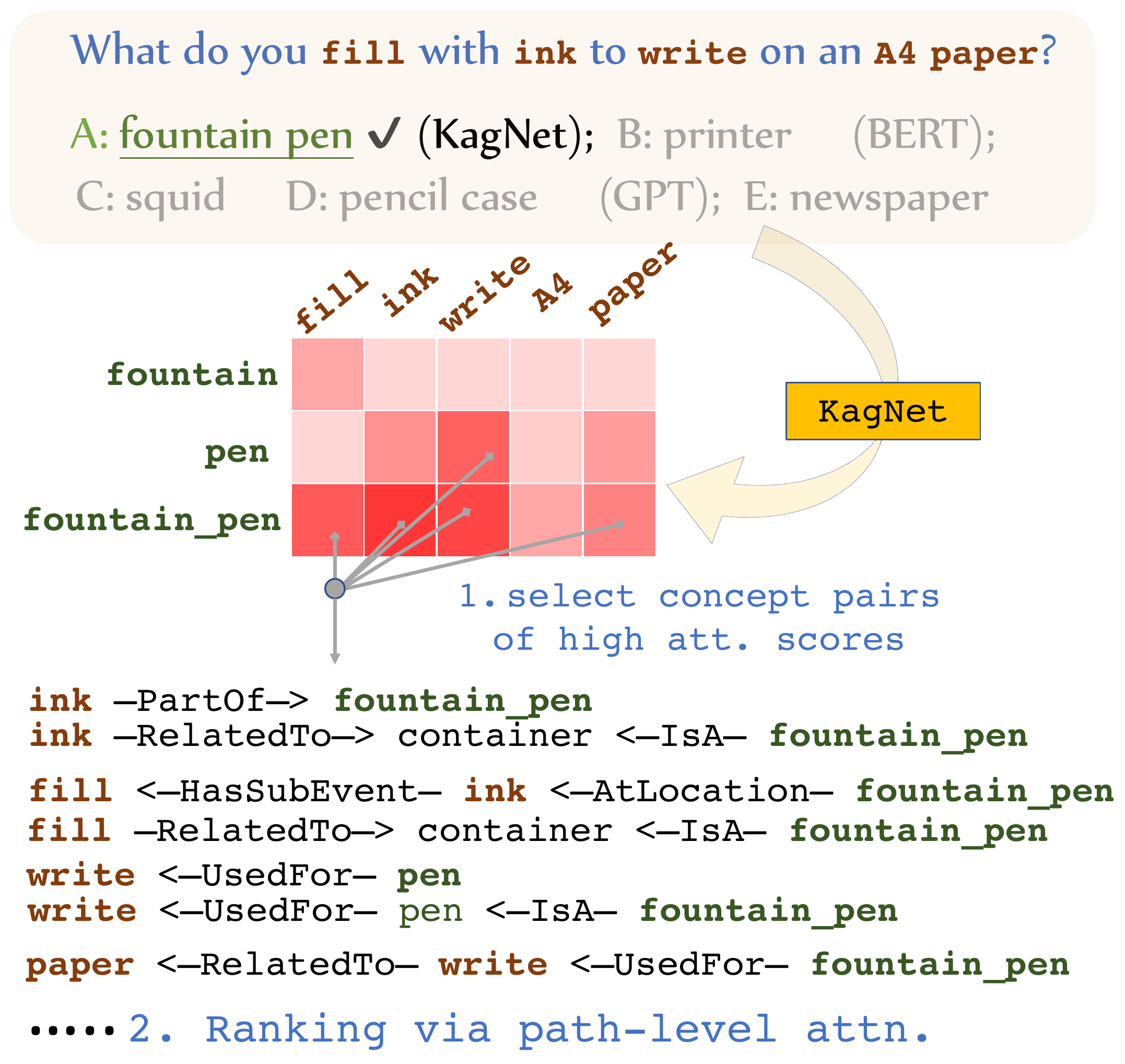}
	\caption{An example of interpreting model behaviors by hierarchical attention scores. }
	\label{fig:interpret}
\end{figure}

Figure~\ref{fig:interpret} shows an example how we can analyze our~\KagNet framework through both pair-level and path-level attention scores.
We first select the concept-pairs with highest attention scores and then look at the (one or two) top-ranked paths for each selected pair.
We find that paths located in this way are highly related to the inference process and also shows that noisy concepts like `fountain' will be diminished while modeling.

\subsection{Model Transferability.} 
We study the transferability of a model that is trained on \texttt{CommonsenseQA} (CSQA) by directly testing it with another task while fixing its parameters.
Recall that we have obtained a \textsc{Bert-Large} model and a \KagNet model trained on CSQA. Now we denoted them as \textsc{Csqa-Bl}~and \textsc{Csqa-Kn} to suggest that they are not trainable anymore. 

In order to investigate their transferability, we separately test them on \texttt{SWAG}~\cite{Zellers2018SWAGAL} and \texttt{WSC} \cite{Levesque2011TheWS} datasets. We first test them the 20k validation examples in \texttt{SWAG}.
\textsc{Csqa-Bl} has an accuracy of $56.53\%$, while our fixed \textsc{Csqa-Kn} model achieves $59.01\%$. 
Similarly, we also test both models on the \texttt{WSC-QA}, which is converted from the \texttt{WSC} pronoun resolution to a multi-choice QA task.

The \textsc{Csqa-BL} achieves an accuracy of $51.23\%$, while our model \textsc{Csqa-KN} scores $53.51\%$. 
These two comparisons further support our assumption that \KagNet, as a knowledge-centric model, is more extensible in commonsense reasoning.  
As we expect for a good knowledge-aware frameworks to behave, our \KagNet indeed enjoys better transferablity than only fine-tuning large language encoders like \textsc{Bert}. 
\subsection{Recent methods on the leaderboard.}
We argue that the \KagNet utilizes the \texttt{ConceptNet} as the only external resource and other methods are improving their performance in orthogonal directions:
1) we find that most of the other recent submissions (as of Aug. 2019) with public information on the leaderboard utilize larger additional textual corpora (e.g. top 10 matched sentences in full Wikipedia via information retrieval tools), and fine-tuning on larger pre-trained encoders, such as \texttt{XLNet}~\cite{Yang2019XLNetGA}, \texttt{RoBERTa}~\cite{Liu2019RoBERTaAR}.
2) there are also models using multi-task learning to transfer knowledge from other reading comprehension datasets, such as \texttt{RACE}~\cite{Lai2017RACELR} and \texttt{OpenBookQA}~\cite{Mihaylov2018CanAS}.

An interesting fact is that the best performance on the OFtest set is still achieved the original fine-tuned \texttt{RoBERTa} model, which is pre-trained with copora much larger than \textsc{Bert}. 
All other \texttt{RoBERTa}-extended methods have negative improvements. 
We also use statement vectors from \texttt{RoBERTa} as the input vectors for \KagNet, and find that the performance on OFdev marginally improves from $77.47\%$ to $77.56\%$. 
Based on our above-mentioned failed cases in error analysis, we believe fine-tuning \texttt{RoBERTa} has achieved the limit due to the annotator biases of the dataset and the lack of comparative reasoning strategies.


\section{Related Work}
\noindent
\textbf{Commonsense knowledge and reasoning.}  
There is a recent surge of novel large-scale datasets for testing machine commonsense with various focuses,
such as situation prediction (\texttt{SWAG})~\cite{Zellers2018SWAGAL}, social behavior understanding~\cite{sap2018atomic, Sap2019SocialIQACR}, visual scene comprehension~\cite{Zellers2019FromRT}, and general commonsense reasoning~\cite{Talmor2018CommonsenseQAAQ},
which encourages the study of supervised learning methods for commonsense reasoning. 
\citeauthor{Trinh2018ASM} (2018) find that large language models show promising results in WSC resolution task~\cite{Levesque2011TheWS}, but this approach can hardly be applied in a more general question answering setting and also not provide explicit knowledge used in inference.
A unique merit of our \KagNet method is that it provides grounded explicit knowledge triples and paths with scores, such that users can better understand and put trust in the behaviors and inferences of the model.

\smallskip
\noindent
\textbf{Injecting external knowledge for NLU.}  
Our work also lies in the general context of using external knowledge to encode sentences or answer questions. 
\citeauthor{Yang2017LeveragingKB} (2017) are the among first ones to propose to encode sentences by keeping retrieving related entities from knowledge bases and then merging their embeddings into LSTM networks computations, to achieve a better performance on entity/event extraction tasks.
\citeauthor{weissenborn2017dynamic} (2017), 
\citeauthor{Mihaylov2018KnowledgeableRE} (2018), and \citeauthor{Annervaz2018LearningBD} (2018) follow this line of works to incorporate the embeddings of related knowledge triples at the word-level and improve the performance of natural language understanding tasks.
In contrast to our work,
they do not explicitly impose graph-structured knowledge into models
, but limit its potential within transforming word embeddings to concept embeddings.

Some other recent attempts \cite{Zhong2018ImprovingQA, Wang2018ImprovingNL} to use ConceptNet graph embeddings are adopted and compared in our experiments (\secref{sec:exp}).
\citeauthor{Rajani2019ExplainYL} (2019) propose to manually collect more human explanations for correct answers as additional supervision for auxiliary training.
\KagNet-based framework focuses on injecting external knowledge as an explicit graph structure, and enjoys the relational reasoning capacity over the graphs.

\smallskip
\noindent
\textbf{Relational reasoning.}  
\KagNet can be seen as a knowledge-augmented \texttt{Relation Network} module (RN)~\cite{Santoro2017ASN}, which is proposed for the visual question answering task requiring relational reasoning (i.e. questions about the relations between multiple 3D-objects in an image). 
We view the concepts in the questions and answers as objects and effectively utilize external knowledge graphs to model their relations from both semantic and symbolic spaces (\secref{sec:lstm}), while prior methods mainly work on the semantic one.







\section{Conclusion}
We propose a knowledge-aware framework for learning to answer commonsense questions. The framework first constructs schema graphs to represent relevant commonsense knowledge, and then model the graphs with our \KagNet module. 
The module is based on a \texttt{GCN-LSTM-HPA} architecture, which effectively represent graphs for relational reasoning purpose in a transparent, interpretable way, yielding a new state-of-the-art results on a large-scale general dataset for testing machine commonsense.
Future directions include better question parsing methods to deal with negation and comparative question answering, as well as incorporating knowledge to visual reasoning. 


%
%
%

\section*{Acknowledgments}
This work has been supported in part by National Science Foundation SMA 18-29268, DARPA MCS and GAILA, IARPA BETTER, Schmidt Family Foundation, Amazon Faculty Award, Google Research Award, Snapchat Gift and JP Morgan AI Research Award. We would like to thank all the collaborators in the INK research lab for their constructive feedback on the work.
\bibliography{emnlp2019}
\bibliographystyle{acl_natbib}

\end{document}